\documentclass{article}

\usepackage[preprint, nonatbib]{neurips_2021}

\usepackage[utf8]{inputenc} % allow utf-8 input
\usepackage[T1]{fontenc}    % use 8-bit T1 fonts
\usepackage{hyperref}       % hyperlinks
\usepackage{url}            % simple URL typesetting
\usepackage{booktabs}       % professional-quality tables
\usepackage{amsfonts}       % blackboard math symbols
\usepackage{nicefrac}       % compact symbols for 1/2, etc.
\usepackage{microtype}      % microtypography
\usepackage{xcolor}         % colors
\usepackage{graphicx}
\usepackage{amsmath}
\usepackage[ruled,lined,commentsnumbered]{algorithm2e}

\title{STR-GODEs: Spatial–Temporal-Ridership Graph ODEs for Metro Ridership Prediction}

\author{%
  Chuyu Huang\\
  Peking University\\
  \texttt{universe\_hcy@pku.edu.cn} \\
}

\begin{document}

\maketitle

\begin{abstract}
The metro ridership prediction has always received extensive attention from governments and researchers. Recent works focus on designing complicated graph convolutional recurrent network architectures to capture spatial and temporal patterns. These works extract the information of spatial dimension well, but the limitation of temporal dimension still exists. We extended Neural ODE algorithms to the graph network and proposed the STR-GODEs network, which can effectively learn spatial, temporal, and ridership correlations without the limitation of dividing data into equal-sized intervals on the timeline. While learning the spatial relations and the temporal correlations, we modify the GODE-RNN cell to obtain the ridership feature and hidden states. Ridership information and its hidden states are added to the GODESolve to reduce the error accumulation caused by long time series in prediction. Extensive experiments on two large-scale datasets demonstrate the efficacy and robustness of our model. \footnote{Our code and benchmarks are available at https://github.com/universe-hcy/STR-GODEs.}

\end{abstract}

\section{Introduction}

With the increasing population of cities, metro lines have become more and more crowded. For improving the service efficiency of the metro system and facilitating passenger travel, metro ridership prediction has become hotspot research in the community of Intelligent Transportation Systems (ITSs).
Recent works focus on designing complicated graph convolutional recurrent network architectures to capture spatial and temporal patterns.
These works depend on dividing the ridership data into equal-sized intervals on the timeline and counting the ridership as the time period. In metro ridership data, there are distinct characteristics of peak and valley periods. The dense ridership during the peak period needs to be divided more finely, while the ridership in the low valley period is less, and there is not enough data when the time interval is small.
This preprocessing destroys information about the ridership over time, which may be information about potential variables. At the same time, the problem is limited to time period and time period prediction, which is not conducive to real-time data collection and feedback to passenger users.

A more practical approach is to construct a continuous-time model that defines a potential state at all times. Chen et al.\cite{chen2018neural} proposed a new family of continuous time-series models (NODE) which directly models the dynamics of network hidden states with an ODE solve. 
Rubanova et al.\cite{rubanova2019latent} generalize state transitions in RNNs to continuous-time dynamics specified by Neural ODEs.and proposes latent ODE network defined in continuous time, which makes it more effective in irregular sampling data. Zhou et al.\cite{zhou2021urban, zhouenhancing} Applied the NODE to the high-resolution reconstruction and prediction of urban flow grid map. These work provide a new direction for metro passenger flow prediction.

Previous work on NODE mainly focused on Euclidean data such as sequences and matrices. We modified it to make it suitable for non-Euclidean data, such as graph networks, called Graph ODEs(GODE).To facilitate performance comparison between models, we used the three complementary graphs proposed by PVCGN\cite{liu2020physical} as spatial information to learn the relationship between subway stations.
Inspired by latent ODEs, we use a GODE-RNN structure as encoder to learn the initial state $Z_{0}$, and hidden states containing the characteristics of ridership. Subsequently, we modify the GODEs network in prediction so that the ridership information and its hidden vectors can be incorporated into the prediction process, which improves the cumulative error caused by the long time series. 
To verify the effectiveness of our STR-GODEs, we conduct experiments on two large-scale benchmarks(i.e., Shanghai Metro and Hangzhou Metro) with widely used baselines and state-of-the-art models and the evaluation results show that our approach outperforms existing state-of-the-art methods under various comparison circumstances. 

In summary, our contribution is threefold:
\begin{itemize}
\item We further extend the neural ODEs network to the time series prediction of graph networks, and to the best of our knowledge, firstly apply it to the metro ridership prediction. 
\item By combining the ridership data and its hidden state, we improve the performance of the prediction model of neural grouph ODEs network so that it can reduce the cumulative error over a long time series.
\item our STR-GODEs has better effect and robustness, and is no longer limited by the same time slice and equal distance prediction in previous work. It can be better applied in real life. Since we have only done some early work, this direction is very prospective.
\end{itemize}

\section{Background and Related Works}

\textbf{Traffic States Prediction} Traffic forecasting depends on the combination of spatial-temporal feature and has been studied for decades.
Traditional methods\cite{liu2019deeppf, milenkovic2018sarima, williams2003modeling, tan2009aggregation} can only consider the relation of a single station in temporal dimension, lack spatial information, and can only learn simple time series models.In recent years, deep neural network has become the mainstream method in this field. The early work \cite{yao2018deep, yao2019revisiting, liu2018attentive} mainly divided the studied cities into regular grid maps, and transformed the raw traffic data into tensors. CNN is used to capture the spatial correlation among near regions. This preprocessing method will bring structural noise information into the traffic system with irregular topology, which will affect the effect of the model.
Graph convolution network\cite{defferrard2016convolutional, kipf2016semi} greatly improves the disadvantages of previous work on spatial information.
Recent works\cite{zhao2019t, geng2019spatiotemporal, bai2020adaptive} focus on designing complicated graph convolutional recurrent network architectures to capture the spatial and temporal patterns.
DCRNN \cite{li2017diffusion} re-formulates the spatial dependency of traffic as a diffusion process and extends the previous GCN to a directed graph.
Graph-WaveNet\cite{wu2019graph} captures the hidden spatial dependency in the data by constructing a adaptive dependency matrix and learning it through node embedding.
Liu et al.\cite{liu2020physical} construct three complementary graphs to utilize the metro physical topology information. A Graph Convolution Gated Recurrent Unit (GC-GRU) is applied to learn the spatial-temporal representation and a Fully-Connected Gated Recurrent Unit (FC-GRU) is applied to capture the global evolution tendency. However, due to the limitation of RNN structure, there are still some deficiencies in the temporal dimension.

\textbf{Graph Neural Network} Unlike traditional Euclidean data, graph data is difficult to be embedded into Euclidean space losslessly. Graph
Convolution Networks (GCN)\cite{defferrard2016convolutional, kipf2016semi} have been proposed to automatically learn feature representation on graphs,which is widely used in node classification\cite{kipf2016semi}, link prediction\cite{zhang2018link}, and graph classification\cite{ying2018hierarchical}.\cite{wu2020comprehensive}
There are two mainstreams of graph convolution networks, the spectralbased
approaches \cite{bruna2013spectral, defferrard2016convolutional, kipf2016semi}and the spatial-based approaches\cite{atwood2016diffusion, gilmer2017neural, hamilton2017inductive}. In these approaches, the adjacency matrix is considered as prior knowledge and is fixed throughout training.
Recently researchers\cite{li2018adaptive, wu2019graph, bai2020adaptive} pay more attention on operating on both spatial and temporal dimensions without the pre-defined graph structure.

\textbf{Neural Ordinary Differential Equations} Built upon previous works\cite{weinan2017proposal, lu2018beyond}, Chen et al.\cite{chen2018neural} proposed a new family of continuous-time models named Neural ODEs,which can be interpreted as a continuous equivalent of Residual networks\cite{he2016deep}.
A basic formulation of Neural ODEs is shown as:
\begin{eqnarray}
\frac{dh(t)}{dt} = f_{\theta}(h(t),t),\quad h(t_{0}) = h_{0}
\end{eqnarray} 
where $f_{\theta}$ is parametrized as a neural network.The hidden state h(t) is defined at all times, and can be evaluated at any desired times using a numerical black-box ODE solver: 
\begin{eqnarray}
h_{0},\ldots,h_{N} = ODESolve(f_{\theta},h_{0},(t_{0},\ldots,t_{N}))
\end{eqnarray} 
By using adjoint method\cite{chen2018neural},the gradients w.r.t.$\theta$ can be computed memory-efficiently during back propagation. Recent works\cite{dupont2019augmented, massaroli2020stable, davis2020time, yan2019robustness, ghosh2020steer} have tried to analyze this framework theoretically, overcome the instability issue and improve memory-efficiency. Researchers also pay attention to apply NODE to other fields such as medical image\cite{pinckaers2019neural}, reinforcement learning\cite{du2020model}, video generation\cite{kanaa2019simple, yildiz2019ode2vae} and graph data\cite{xhonneux2020continuous}.

\section{Method}

\subsection{Problem Definition}

Previous work did not support predictions over continuous or unequal density time intervals. To facilitate performance comparison between models, we follow the traditional prediction framework.
Assuming N is the number of Metro stations, the time period T data can be expressed as $X_{t} = (X_{t}^{1},X_{t}^{2},...,X_{t}^{N})$, where $X_{t}^{i} \in \mathbb{R}^{2}$ represents the the passenger counts of inflow/outflow of the station i at time interval t.
our target is to predict the future ridership sequence based on the observed historical values:
\begin{eqnarray} 
\hat X_{t + 1},\hat X_{t + 2},...,\hat X_{t + m} = STR-GODEs(X_{t-n+1},X_{t-n+2},...,X_{t})
\end{eqnarray} 

\subsection{STR-GODEs}

Suppose $Z(t)$ is the latent state including the ridership feature at time interval t, and $f(t,Z(t),\theta)$ is the function of ridership changing with spatial,temporal and ridership information. By Euler method,  
there is $Z(t + \Delta t) = Z(t) + \Delta t * f(t,Z(t),\theta)$.

The primal problem can be abstracted as we learn the ridership trend function $f(t,Z(t),\theta)$ according to the training data and pre-defined graph (Physical-Virtual Graph\cite{liu2020physical}), obtain the initial state $Z_{0}$ through the observation sequence, and predict the subsequent sequence.

Chen et al.\cite{chen2018neural}  present a continuous-time, generative approach to modeling time series. Each trajectory is determined from a local initial state, $Z_{t_{0}}$ , and a global set of latent dynamics shared across all time series:
\begin{gather}
Z_{t_{0}} \sim{p(Z_{t_{0}})} \\
Z_{t_{0}}, Z_{t_{1}},\ldots,Z_{t_{n}} = ODESolve(Z_{t_{0}},f,\theta_{f},t_{0},\ldots,t_{N})  \\
each\quad x_{t_{i}} \sim{p(x|Z_{t_{i}},\theta_{x})} 
\end{gather}
We follow the framework of Chen et al.\cite{chen2018neural} for both training and prediction.
We modify the original ODE Block to make it suitable for graph network data, called GODE Block. 
As shown in figure 1, inspired by Rubanova et al.\cite{rubanova2019latent}, we use the GODE-RNN structure to obtain the local initial state $Z_{t_{0}}$ by learning the spatial,temporal and ridership feature of $X_{t},X_{t + 1},...,X_{t + n}$.
In particular, due to the need to gather information from adjacent metro stations in GODE Block, we use GODE-RNN to learn the approximate posterior of local initial state $Z_{t_{0}}$ directly instead of learning the mean and standard deviation.
\begin{gather}
q(z_{t_{0}}|\{x_{i},t_{i}\}^{N}_{i=0}) = Z_{t_{0}}\quad where Z_{t_{0}},hidden_{t_{0}} = g(ODE-RNN_{\phi}(x_{i},t_{i})^{N}_{i=0})
\end{gather}

Furthermore, by combining the ridership information and its hidden state, we further improve the performance of the prediction model so that it can reduce the cumulative error over a long time series.

\begin{figure}[h]
  \centering
  \includegraphics[width=\linewidth]{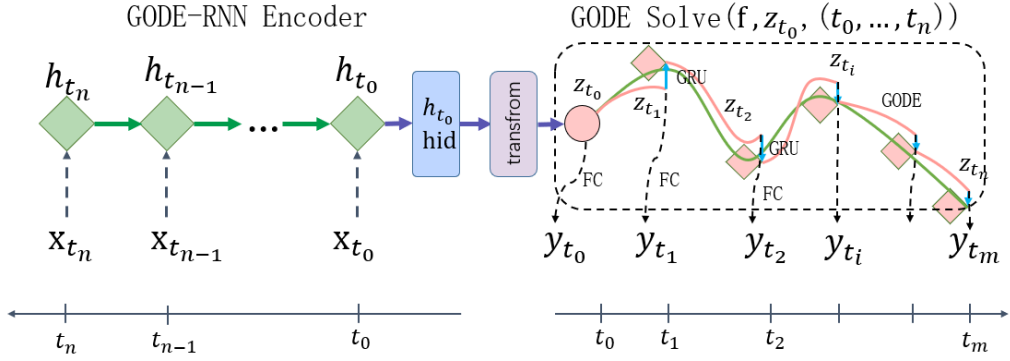}
  \caption{The illustration of encoder and decoder used in STR-GODEs.In GODESolve, the green line represents ground truth, and the pink line represents the predicted result obtained by GODE. At some special time points, we correct the predicted result by introducing ridership information and its hidden state through GRU cell.}
  \label{fig:example}
\end{figure}

\subsection{Physical-Virtual Graphs}

In this section, we follows the pre-defined graph proposed by PVCGN\cite{liu2020physical} to learn spatial information. In our work, the physical graph, similarity graph and correlation graph are denoted as $\mathcal{G}_{p}=(\mathcal{V},\mathcal{E}_{p},W_{p})$, $\mathcal{G}_{s}=(\mathcal{V},\mathcal{E}_{s},W_{s})$ and $\mathcal{G}_{c}=(\mathcal{V},\mathcal{E}_{c},W_{c})$, where $\mathcal{V}$ is a set of nodes represent real-world metro stations($|\mathcal{V}| = N$), $\mathcal{E}$ is a set of edges and $W \in \mathcal{R}^{N\times N}$ denotes the weights of all edges.

\textbf{Physical Graph}:$\mathcal{G}_{p}$is directly built based on the
physical topology of the studied metro system. We first construct a physical connection matrix $P \in \mathbb{R}^{N\times N}$, where $P(i,j)=1$ if
there exists an edge between node i and j, or else $P(i,j)=0$. Note that each
diagonal value P(i, i) is directly set to 0.Finally, the edge weight $W_{p}$ is obtained by performing a linear normalization on each row:
\begin{eqnarray} 
W_{p}(i,j) = \frac{P(i,j)}{\sum_{k=1}^{N}P(i,k)}
\end{eqnarray} 
\textbf{Similarity Graph}:The similarities of metro stations are used to guide the construction of $\mathcal{G}_{s}$. We construct a similarity score matrix $S \in \mathbb{R}^{N\times N}$, where the score S(i, j) between station i and
j is computed with Dynamic Time Warping\cite{berndt1994using}:
$ S(i,j) = exp(-DWT(X^{i},X^{j})) $.
Based on the matrix S, we select some station pairs to build edges $\mathcal{E}_{s}$ with a predefined similarity threshold 0.1 for HZMetro and by choosing the top ten stations with high similarity scores for SHMetro.
Finally, we calculate the edge weights $W_{s}$ by conducting row normalization on S:
\begin{eqnarray} 
W_{s}(i,j) = \frac{S(i,j)}{\sum_{k=1}^{N}S(i,k)\cdot L(E_{s},i,k)}
\end{eqnarray} 
where $L(\mathcal{E}_{s},i,k) = 1$ if $\mathcal{E}_{s}$ contains an edge connecting node i and k, or else $L(\mathcal{E}_{s},i,k) = 0$.

\textbf{Correlation Graph}: We utilize the origin-destination distribution of ridership to build the virtual graph $\mathcal{G}_{c}$. First, we construct a correlation ratio matrix $C \in \mathbb{R}^{N\times N}$:
$C(i,j) = \frac{D(i,j)}{\sum_{k=1}^{N}D(i,k)}$
where D(i, j) is the total number of passengers that traveled from station j to station i in the whole training set.
Based on the matrix C, we select some station pairs to build edges $\mathcal{E}_{c}$ with a predefined similarity threshold 0.02 for HZMetro and by choosing the top ten stations with high similarity scores for SHMetro.
Then, the edge weights $W_{c}$ is calculated by:
\begin{eqnarray} 
W_{c}(i,j) = \frac{C(i,j)}{\sum_{k=1}^{N}C(i,k)\cdot L(E_{c},i,k)}
\end{eqnarray}

\subsection{GODE Block}

In general, the graph convolutional networks is a method to aggregate the information of neighbor nodes around the target node and can be expressed as\cite{defferrard2016convolutional, kipf2016semi, hamilton2017inductive, velivckovic2017graph}:
\begin{eqnarray}
x_{i}^{k} = \gamma(x_{i}^{k-1},\square_{j \in \aleph (i)}\phi^{k}(x_{i}^{k-1},x_{j}^{k-1},e_{i,j}))
\end{eqnarray}
where $x_{i}^{k}$ refers to the hidden state of node i in the k-th layer, $e_{i,j}$ refers to the eigenvector of the edge from node i to node j, $\square$ is the aggregation method of neighbor information, $\phi$ is the transformation method of neighbor information and $\gamma$ is a transformation that fuses one's own feature with its aggregated neighbors'information.

These and related methods can be understood as special cases of
a simple differentiable message-passing framework (Gilmer et al. 2017\cite{gilmer2017neural}):
\begin{eqnarray}
h_{i}^{l+1}=\sigma \left(\sum\limits_{m\in\mathcal{M_{i}}}g_{m}(h_{i}^{l},h_{j}^{l})\right)
\end{eqnarray}
where $h_{i}^{l+1} \in \mathcal{R}^{d}$ is hidden state of node $v_{i}$ in the
l-th layer of the neural network. Incoming messages of the form $g_{m}(\cdot,\cdot)$ are accumulated and passed through an element-wise activation function$\sigma(\cdot)$. $\mathcal{M}_{i}$denotes the set of incoming
messages for node $v_{i}$ and is often chosen to be identical to
the set of incoming edges.
It can also be expressed in the following form: 
\begin{eqnarray}
h_{t+1}=h_{t}+f(h_{t},\theta_{t},\mathcal{G})
\end{eqnarray}

According to Neural ODEs\cite{chen2018neural}, the hidden state h(t) can be defined as the solution to an ODE initial-value problem: 
\begin{eqnarray}
\frac{dh(t)}{dt} = f_{\theta}(h(t),t,\mathcal{G}),where\quad h(t_{0}) = h_{0}
\end{eqnarray}

Inspired by R-GCN\cite{schlichtkrull2018modeling}, we define our GODE-function by using learnable weights $\mathbf{\Theta}_0$ to learn the changes of stations latent states over time, and learnable weights $\mathbf{\Theta}_r$ to learn the association relationship between stations. For every station,there are: 
\begin{eqnarray}
\mathbf{z}^{t+\Delta t}_i = relu(\mathbf{\Theta}_0 \cdot \mathbf{z}^{t}_i +
        \sum_{r \in \mathcal{R}} \sum_{j \in \mathcal{N}_r(i)}
        \frac{1}{|\mathcal{N}_r(i)|} \mathbf{\Theta}_r \cdot \mathbf{z}^{t}_j)
\end{eqnarray}

For any time $t_{i}$, we can calculate the latent representation $Z_{t_{i}}$ by:
\begin{eqnarray}
Z_{t_{i}} = Z_{t_{i-1}} + \int_{t_{i-1}}^{t_{i}} {f_{\theta}(h(t),t,\mathcal{G})} \,{\rm d}t
\end{eqnarray}

\begin{algorithm}[h]
\caption{The GODE-RNN. The difference is highlighted in \textcolor{blue}{blue}}\label{algorithm}
\quad \textbf{Input:} Data points and their timestamps$\{(x_{i},t_{i})\}_{i=1...N}$

\quad$h_{N},hidden = 0, 0$

\quad \textbf{for} i in N,N-1,...,1 \textbf{do}

\quad\quad  $time\_points = utils.linspace\_vector(t_{i-1}, t_{i}, n\_intermediate\_tp)$

\quad\quad $h^{\prime} = \textcolor{blue}{GODESolve}(f_{\theta},h_{i-1},time\_points)$

\quad\quad $h_{i},\textcolor{blue}{hidden} = RNNCell(h^{\prime}, \textcolor{blue}{hidden}, x_{i})$

\quad \textbf{end for}

\quad \textbf{Return $h_{0},hidden$}

\end{algorithm}

\subsection{STR-GODEs encoder}

Rubanova et al. proposed using an ODE-RNN as the encoder for a latent ODEs model to learn the mean $\mu_{z_{0}}$ and standard deviation $\sigma_{z_{0}}$ of the approximate posterior $q(z_{t_{0}}|\{x_{i},t_{i}\}^{N}_{i=0})$ and the local initial state $Z_{t_{0}} = \mathcal{N}(\mu_{z_{0}}, \sigma_{z_{0}})$.

As shown in algorithm 1, we made some modifications on ODE-RNN. 
We replaced ODESolve with GODESolve, which can be applied to graph data.
We use GODE-RNN to learn the approximate posterior of local initial state $Z_{t_{0}}$ directly instead of learning the mean and standard deviation. The hidden state of ridership information is learned and stored in hidden, which is combined with $Z_{i}$ in RNNCell. In our work, we use the classical GRU\cite{cho2014learning} structure as the RNNCell. Specifically, the reset gate 
$r_{t} = \{r_{t}^{1},r_{t}^{2},\ldots,r_{t}^{N} \} $, update gate $z_{t} = \{z_{t}^{1},z_{t}^{2},\ldots,z_{t}^{N} \}$, new information $N_{t} = \{N_{t}^{1},N_{t}^{2},\ldots,N_{t}^{N} \}$ are computed by:
\begin{gather}
r_{t} = \sigma(W_{r} \cdot [Z_{t - 1},h_{t - 1},x_{t}])\\
z_{t} = \sigma(W_{z} \cdot [Z_{t - 1},h_{t - 1},x_{t}])\\
N_{t} = tanh(W_{N}\cdot[r_{i}*h_{t-1},x_{t}])\\
h_{t} = (1 - z_{t})* N_{t-1} + z_{t} * h_{t - 1}\\
Z_{t} = (1 - z_{t})* N_{t-1} + z_{t} * Z_{t - 1}
\end{gather}

\begin{algorithm}[h]
\caption{The GODESolver\_prediction. }
\label{algorithm}
\quad \textbf{Input:} prediction timestamps$\{t_{i}\}_{i=1...M},Z_{0}$,hidden,Data points$\{x_{i}\}_{i=1...N}$

\quad \textbf{for} i in 1,2,...,M \textbf{do}

\quad\quad $Z^{\prime} = odeint(f_{\theta},Z_{i-1},(t_{i-1},t_{i}))$

\quad\quad $y_{i} = output\_layer(Z^{\prime})$

\quad\quad $x_{i} = x_{i}$ if $x_{i}$ is available Data points else $y_{i}$

\quad\quad $Z_{i}, hidden = RNNCell(Z^{\prime}, hidden, x_{i})$

\quad \textbf{end for}

\quad \textbf{Return $\{y_{i}\}_{i=1...M}$}

\end{algorithm}

\subsection{STR-GODEs decoder}

Before prediction, we convert the latent state $h_{0},hidden$ obtained from GODE-RNN to get $Z_{0},hidden$ with a transform block which is composed of a fully-connected layer, a tanh activation function, and a fully-connected layer. 

Previous work uses Neural ODEs as decoder for prediction and when the sequence is long, the error will accumulate gradually. Similar to the GODE-RNN, 
We introduce ridership information and its hidden state into GODESolver through RNNCell structure, as shown in algorithm 2.
The experimental results show that it can reduce the cumulative error over a long time series. Specifically,  we use the classical GRU\cite{cho2014learning} structure described in session 3.5 as the RNNCell.

\section{Experiment}

\subsection{Datasets}

To evaluate the performance of our work, we conduct experiments on two real-world traffic: SHMetro and HZMetro\footnote{https://tianchi.aliyun.com/competition/entrance/231708/information}.   

\textbf{SHMetro}: The SHMetro dataset refers to the ridership data on the metro system of Shanghai, China. There are 288 metro stations connected
by 958 physical edges. A total of 811.8 million transaction
records were collected from Jul. 1st 2016 to Sept. 30th 2016,
with 8.82 million ridership per day. Following the setting of PVCGN\cite{liu2020physical}, for each station, we measured its inflow and outflow every 15 minutes by counting the number of passengers entering or exiting the station. The ridership data of the first two months and that of the last three weeks are used for training and testing, while the ridership data of the remaining days are used for validation.

\textbf{HZMetro}: The HZMetro dataset refers to the ridership data on the metro system of Hangzhou, China. There are 80 metro stations connected
by 248 physical edges with 2.35 million ridership per day. The time interval of this dataset is also set to 15 minutes. Similar to SHMetro, this dataset is divided into three parts, including a training set (Jan. 1st - Jan. 18th), a validation set (Jan. 19th - Jan. 20th), and a testing set (Jan. 21th - Jan. 25th).

\subsection{Experimental Settings}

We have trained the official code of PVCGN\cite{liu2020physical}, which has a small gap with the results of corresponding papers.
To reduce the effect difference caused by different parameters and facilitate the model performance comparison, we keep the parameters in STR-GODEs the same as those in the official code of PVCGN.

We implement our STR-GODEs in Python with Pytorch 1.6.0 and executed on a server with one NVIDIA Titan X GPU card. 
The lengths of input and output sequences are set to 4 simultaneously. The input data and the ground-truth of output are normalized with Z-score Normalization\footnote{https://en.wikipedia.org/wiki/Standard\_score}
before being fed into the network. 
The batch size is set to 8 and the feature dimensionality d is set to 256.
The initial learning rate is 0.001 and its decay ratio is 0.1.
We optimize the models by Adam\cite{kingma2014adam} optimizer for a maximum of 200 epochs by minimizing the mean absolute error between the predicted results and the corresponding ground-truths.
The best parameters for all deep learning models are chosen through a carefully parameter-tuning process on the validation set.

\subsection{Compared Methods}
To evaluate the overall performance of our work, we compare STR-GODEs with widely used baselines and state-of-the-art models:

\textbf{DCRNN}: Huang et al.\cite{huang2019diffusion} proposed a deep learning framework specially designed for traffic prediction, which uses bidirectional random walks
on graphs to capture spatial dependencies and an encoder-decoder architecture  to learn temporal correlation. We implement this method with its official code.

\textbf{Graph-WaveNet}: Wu et al.\cite{wu2019graph} developed an adaptive dependency matrix to capture the hidden spatial dependencies, and used a stacked dilated 1D convolution component to handle very long sequences. We implement this method with its official code.

\textbf{ST—ODEs}: Zhou et al. applied the NODE to the prediction of urban flow grid maps. They concats the data over different time intervals, feeds it into CNN layer to get the initial state $Z_{0}$, and then uses the neural ODEs network for prediction.In this experiment, we reproduced the code and extended the ODE network to GODE network which can be used for graph data.

\textbf{latent ODEs}: Rubanova et al. refine the Latent ODE model of Chen et al. \cite{chen2018neural} by using the ODE-RNN as a recognition network, where ODE-RNN is generalized from RNNs with continuous-time hidden dynamics defined by ordinary differential.
In the experiment, we modify its official code and extended the ODE network to GODE network which can be used for graph data.

\textbf{PVCGN}: Liu et al.\cite{liu2020physical} developed a
Seq2Seq model with GC-GRU and FC-GRU to forecast the
future metro ridership sequentially where Graph Convolution Gated Recurrent Unit (GC-GRU) is applied to learn the spatial-temporal representation and Fully-Connected Gated Recurrent Unit (FC-GRU) is applied to capture the global evolution tendency. We implement this method with its official code.

\begin{table}[h] %开始一个表格environment，表格的位置是h,here。
\centering
\begin{tabular}{cccccccc} %设置了每一列的宽度，强制转换。
\hline
Time & Metric & DCRNN &Graph-wavenet & ST—ODEs & latent ODEs & PVCGN & STR-GODEs  \\ %用&来分隔单元格的内容 \\表示进入下一行
\hline %画一个横线，下面的就都是一样了，这里一共有4行内容
\hline
      & MAE  & 23.81   & 24.01   & 24.54    & 22.79   & \textbf{22.70}   & 22.87\\
15min & RMSE & 40.52   & 40.68   & 42.43    & 38.61   & 37.78   & \textbf{37.71}\\
      & MAPE & 14.13\% & 14.31\% & 15.01\%  & \textbf{13.54}\% & \textbf{13.54}\% & 13.78 \%\\
\hline
      & MAE  & 25.20   & 25.45   & 25.34    & 23.53   & 23.71   & \textbf{23.24}\\
30min & RMSE & 42.51   & 42.02   & 43.49    & 39.63   & 39.96   & \textbf{38.69}\\
      & MAPE & 14.89\% & 15.18\% & 15.43\%  & 13.85\% & 14.06\% & \textbf{14.03} \%\\
\hline
      & MAE  & 26.86   & 27.21   & 26.56    & 24.44   & 24.53   & \textbf{23.79}\\
45min & RMSE & 45.93   & 46.03   & 46.00    & 41.22   & 41.31   & \textbf{39.68}\\
      & MAPE & 16.02\% & 17.39\% & 16.50\%  & 14.67\% & \textbf{14.66}\% & \textbf{14.66} \%\\
\hline
      & MAE  & 28.41   & 29.21   & 27.45    & 24.99   & 24.88   & \textbf{24.35}\\
60min & RMSE & 49.23   & 49.77   & 47.61    & 42.20   & 41.61   & \textbf{40.83}\\
      & MAPE & 18.06\% & 19.41\% & 19.01\%  & 16.05\% & 15.68\% & \textbf{15.56} \%\\
\hline
\end{tabular}
\caption{results of conventional prediction experiment on HZMetro} %显示表格的标题
\end{table}

\begin{table}[h] %开始一个表格environment，表格的位置是h,here。
\centering
\begin{tabular}{cccccccc} %设置了每一列的宽度，强制转换。
\hline
Time & Metric & DCRNN &Graph-wavenet & ST—ODEs & latent ODEs & PVCGN & STR-GODEs  \\ %用&来分隔单元格的内容 \\表示进入下一行
\hline %画一个横线，下面的就都是一样了，这里一共有4行内容
\hline
      & MAE  & 24.09   & 24.86   & 24.79    & 23.51   & 23.41   & \textbf{23.21}\\
15min & RMSE & 46.11   & 46.82   & 46.96    & 44.64   & 45.31   & \textbf{44.58}\\
      & MAPE & 17.87\% & 19.98\% & 18.56\%  & 17.64\% & 17.04\% & \textbf{16.99} \%\\
\hline
      & MAE  & 25.23   & 26.73   & 25.77    & 25.06   & 24.36   & \textbf{23.63}\\
30min & RMSE & 49.92   & 52.01   & 50.08    & 48.91   & 48.12   & \textbf{46.28}\\
      & MAPE & 18.32\% & 20.42\% & 19.29\%  & 18.37\% & 17.35\% & \textbf{17.12} \%\\
\hline
      & MAE  & 26.84   & 28.71   & 26.80    & 26.18   & 25.29   & \textbf{24.65}\\
45min & RMSE & 55.07   & 58.32   & 53.59    & 52.60   & 51.42   & \textbf{49.93}\\
      & MAPE & 19.37\% & 21.93\% & 20.59\%  & 19.22\% & 17.84\% & \textbf{17.58} \%\\
\hline
      & MAE  & 28.22   & 30.61   & 27.63    & 26.83   & 26.21   & \textbf{25.56}\\
60min & RMSE & 60.18   & 64.02   & 56.04    & 55.78   & 55.17   & \textbf{53.39}\\
      & MAPE & 20.57\% & 24.22\% & 22.38\%  & 20.28\% & 18.83\% & \textbf{18.25} \%\\
\hline
\end{tabular}
\caption{results of conventional prediction experiment on SHMetro} %显示表格的标题
\end{table}

\subsection{Performance Comparison and Analysis}

We measure the performance of predictive models with 
three widely used metrics - Mean Absolute Error (MAE), Root Mean Square Error (RMSE), and Mean Absolute Percentage Error (MAPE).

\textbf{Conventional prediction experiment}: Table 1 and Table 2 present the prediction performances in conventional prediction experiments of our STR-GODEs and representative comparison methods in HZMetro and SHMetro datasets.
Compared with latent ODE, We can observe that the GODE network combined with ridership data and hidden states can indeed reduce the cumulative error.
Compared with all other methods, STR-GODEs has achieved some improvement in most metrics.

Further more, we extract the prediction of ridership in peak periods(7:30-9:30 and 17:30-19:30) in conventional experiments, as shown in table 6 and table 7.
We can observe that our STR-GODEs outperforms all comparative methods consistently on most metrics on both datasets.
On HZMetro, we achieve a relative improvement of 4.13\% in MAE metric compared with other optimal algorithms on average, 3.59\% in RMSE and 4.28\% MAPE.
On SHMetro, we achieve a relative improvement of 4.32\% in MAE metric compared with other optimal algorithms on average, 4.57\% in RMSE and 1.84\% MAPE.

\begin{table}[h] %开始一个表格environment，表格的位置是h,here。
\centering
\begin{tabular}{cccccccc} %设置了每一列的宽度，强制转换。
\hline
Time & Metric & DCRNN &Graph-wavenet & ST—ODEs & latent ODEs & PVCGN & STR-GODEs  \\ %用&来分隔单元格的内容 \\表示进入下一行
\hline %画一个横线，下面的就都是一样了，这里一共有4行内容
\hline
      & MAE  & 34.22   & 36.42   & 34.89    & 33.17   & 32.68   & \textbf{31.68}\\
15min & RMSE & 53.09   & 56.13   & 54.71    & 52.20   & 50.23   & \textbf{48.78}\\
      & MAPE & 10.02\% & 10.41\% & 10.02\%  & 9.65\% & 9.67\% & \textbf{9.31} \%\\
\hline
      & MAE  & 36.79   & 38.47   & 36.37    & 34.64   & 33.83   & \textbf{32.37}\\
30min & RMSE & 57.13   & 59.05   & 56.84    & 53.82   & 52.59   & \textbf{50.33}\\
      & MAPE & 10.27\% & 10.74\% & 10.13\%  & 9.89\% & 9.58\% & \textbf{9.18} \%\\
\hline
      & MAE  & 37.53   & 38.71    & 36.15    & 34.22   & 33.85   & \textbf{32.06}\\
45min & RMSE & 59.89   & 59.92   & 57.11    & 53.89   & 53.28   & \textbf{50.69}\\
      & MAPE & 10.71\% & 11.63\% & 10.78\%  & 10.44\% & 10.04\% & \textbf{9.62} \%\\
\hline
      & MAE  & 37.16   & 37.89   & 34.85    & 32.71   & 32.05   & \textbf{30.82}\\
60min & RMSE & 58.93   & 60.13   & 54.98    & 52.36   & 51.64   & \textbf{50.45}\\
      & MAPE & 11.13\% & 12.04\% & 11.86\%  & 11.20\% & 10.74\% & \textbf{10.20} \%\\
\hline
\end{tabular}
\caption{results of conventional prediction experiment in peak periods on HZMetro. The peak periods refer to 7:30-9:30 and 17:30-19:30} %显示表格的标题
\end{table}

\begin{table}[h] %开始一个表格environment，表格的位置是h,here。
\centering
\begin{tabular}{cccccccc} %设置了每一列的宽度，强制转换。
\hline
Time & Metric & DCRNN &Graph-wavenet & ST—ODEs & latent ODEs & PVCGN & STR-GODEs  \\ %用&来分隔单元格的内容 \\表示进入下一行
\hline %画一个横线，下面的就都是一样了，这里一共有4行内容
\hline
      & MAE  & 38.12   & 38.84   & 38.32    & 35.31   & 36.57   & \textbf{35.17}\\
15min & RMSE & 67.74   & 68.01   & 66.89    & \textbf{61.87}   & 64.87   & 61.93\\
      & MAPE & 14.07\% & 14.23\% & 13.99\%  & 13.27\% & 13.32\% & \textbf{13.03} \%\\
\hline
      & MAE  & 39.72   & 42.73   & 40.46    & 38.12   & 37.93   & \textbf{35.81}\\
30min & RMSE & 73.14   & 77.48   & 72.20    & 69.02   & 68.91   & \textbf{64.34}\\
      & MAPE & 14.19\% & 14.92\% & 14.57\%  & 13.94\% & 13.49\% & \textbf{13.20} \%\\
\hline
      & MAE  & 41.52   & 45.33     & 41.71    & 38.44   & 38.51   & \textbf{36.21}\\
45min & RMSE & 79.03   & 85.14   & 77.12    & 71.12   & 73.55   & \textbf{67.16}\\
      & MAPE & 15.13\% & 15.79\% & 15.53\%  & 14.67\% & 14.11\% & \textbf{13.78} \%\\
\hline
      & MAE  & 41.13   & 45.74   & 41.10    & 37.51   & 37.87   & \textbf{35.44}\\
60min & RMSE & 78.81   & 89.21   & 78.73    & 71.43   & 73.13   & \textbf{66.99}\\
      & MAPE & 16.12\% & 17.21\% & 16.92\%  & 15.89\% & 15.03\% & \textbf{14.87} \%\\
\hline
\end{tabular}
\caption{results of conventional prediction experiment in peak periods on SHMetro. The peak and valley periods refer to 7:30-9:30 and 17:30-19:30} %显示表格的标题
\end{table}

\textbf{Irregular prediction experiment}: In order to compare the performance of different methods in continuous time or unequal interval time, we use a compromise experiment that the traditional method can also work.

In this experiment, assuming that $X_{t_{1}},X_{t_{2}},\ldots,X_{t_{N}}$ are the observed ridership sequence, where $t_{1},t_{2},\ldots,t_{N}$ is the random subsequence of $t - n + 1,t - n + 2,\ldots,t + 1,\ldots,t + m$, our goal is to predict a the ridership sequence: 
\begin{eqnarray} 
\hat X_{t_{n + 1}},\hat X_{t_{n + 2}},...,\hat X_{t_{n + m}} = STR-GODEs(X_{t_{1}},X_{t_{2}},...,X_{t_{n}})
\end{eqnarray} 

\begin{table}[h] %开始一个表格environment，表格的位置是h,here。
\centering
\begin{tabular}{cccccccc} %设置了每一列的宽度，强制转换。
\hline
Time & Metric & ST—ODEs & latent ODEs & PVCGN & STR-GODEs  \\ %用&来分隔单元格的内容 \\表示进入下一行
\hline %画一个横线，下面的就都是一样了，这里一共有4行内容
\hline
            & MAE  & 31.53   & 19.95   & 22.19    & \textbf{17.70}   \\
$t_{n + 1}$ & RMSE & 63.46   & 40.12   & 38.91    & \textbf{37.31}   \\
            & MAPE & 16.34\% & 11.42\% & 13.12\%  & \textbf{10.14}\% \\
\hline
            & MAE  & 29.70   & 20.83   & 23.57    & \textbf{18.25}   \\
$t_{n + 2}$ & RMSE & 56.69   & 40.21   & 41.22    & \textbf{36.04}   \\
            & MAPE & 15.90\% & 12.34\% & 13.98\%  & \textbf{10.75}\% \\
\hline
            & MAE  & 30.04   & 21.10   & 22.22    & \textbf{18.19}   \\
$t_{n + 3}$ & RMSE & 57.51   & 42.67   & 40.32    & \textbf{37.52}   \\
            & MAPE & 15.82\% & 11.92\% & 13.21\%  & \textbf{10.08}\% \\
\hline
            & MAE  & 33.24   & 22.02   & 20.63    & \textbf{18.45}   \\
$t_{n + 4}$ & RMSE & 68.08   & 44.71   & 38.18    & \textbf{37.41}   \\
            & MAPE & 18.54\% & 13.13\% & 13.08\%  & \textbf{10.74}\% \\
\hline
\end{tabular}
\caption{results of irregular prediction experiment on HZMetro} %显示表格的标题
\end{table}

\begin{table}[h] %开始一个表格environment，表格的位置是h,here。
\centering
\begin{tabular}{cccccccc} %设置了每一列的宽度，强制转换。
\hline
Time & Metric & ST—ODEs & latent ODEs & PVCGN & STR-GODEs  \\ %用&来分隔单元格的内容 \\表示进入下一行
\hline %画一个横线，下面的就都是一样了，这里一共有4行内容
\hline
            & MAE  & 27.32   & 16.61   & 25.18    & \textbf{16.11}   \\
$t_{n + 1}$ & RMSE & 59.35   & \textbf{38.46}   & 51.60    & 38.66   \\
            & MAPE & 19.32\% & 12.55\% & 17.61\%  & \textbf{11.94}\% \\
\hline
            & MAE  & 27.55   & 16.36   & 25.36    & \textbf{15.69}   \\
$t_{n + 2}$ & RMSE & 59.45   & 38.39   & 52.73    & \textbf{37.36}   \\
            & MAPE & 19.99\% & 12.66\% & 18.00\%  & \textbf{11.96}\% \\
\hline
            & MAE  & 27.94   & 15.63   & 25.01    & \textbf{14.96}   \\
$t_{n + 3}$ & RMSE & 61.24   & 37.24   & 53.52    & \textbf{36.33}   \\
            & MAPE & 21.13\% & 12.21\% & 18.21\%  & \textbf{11.19}\% \\
\hline
            & MAE  & 28.93   & 16.12   & 22.29    & \textbf{15.66}   \\
$t_{n + 4}$ & RMSE & 61.88   & 39.88   & 50.83    & \textbf{39.72}   \\
            & MAPE & 24.08\% & 12.24\% & 16.81\%  & \textbf{11.02}\% \\
\hline
\end{tabular}
\caption{results of irregular prediction experiment on SHMetro} %显示表格的标题
\end{table}

Table 3 and Table 4 present the prediction performances in irregular prediction experiments of our STR-GODEs and representative comparison methods in HZMetro and SHMetro datasets. 
Through the experimental results, we can see that Neural GODEs network is superior to the traditional method in irregular prediction and 
compared with other work,  the superiority in effectiveness of the proposed STR-GODEs is verified. 

\section{Conclusion}

In this paper, we extended Neural ODE algorithms to the graph network and proposed the STR-GODEs network, which can effectively learn spatial, temporal, and ridership correlations without dividing data into equal-sized intervals on the timeline. Specifically, we use a GODE-RNN structure as encoder to learn the initial state $Z_{0}$, and hidden states containing the characteristics of ridership. Subsequently, GODE network which can incorporate the ridership information and its hidden vectors is used to predict the future sequence.
Experimental results on two public large-scale datasets demonstrate the superior performance of our algorithm.

In future works, we will make further use of the continuous time series-based characteristics of the model to meet the requirements of real-time prediction and apply the model to the passenger. 
At the same time, we will further optimize the time and space efficiency of the model to enable it to run on mobile devices such as mobile phones.

\medskip

{\small

\bibliographystyle{ieee_fullname}
\bibliography{reference}
}

\end{document}